\documentclass{article}

\usepackage{microtype}
\usepackage{graphicx}
\usepackage{subfigure}
\usepackage{booktabs} 

\usepackage{times}
\usepackage{epsfig}
\usepackage{graphicx}
\usepackage{amsmath}
\usepackage{amssymb}
\usepackage{xcolor}
\usepackage{tabularx}
\usepackage{siunitx}
\usepackage{bbm}
\usepackage{capt-of}
\usepackage{adjustbox}
\usepackage{wrapfig}
\usepackage{algorithmic}
\usepackage{algorithm}
\usepackage{soul}
\usepackage{enumitem}
\DeclareMathOperator*{\argmin}{arg\,min}

\usepackage{hyperref}

\usepackage{pythontex}

\newcommand{\ignoretext}[2]{\pyc{%
  ret = r"#2".replace(r"#1", r'');%
  print(ret);%
}}

\newif\ifdrafting
\draftingtrue 
\ifdrafting
    \newcommand{\hila}[2][]{%
    \ifthenelse{ \equal{#1}{} }
        {\textcolor{teal}{(Hila) #2}}
        {\textcolor{teal}{(Hila) \sout{#1}\xspace{}#2}}
        {\ignoretext{teal}{\sout{#1}\xspace{}#2}}
}

\newcommand{\tali}[2][]{%
    \ifthenelse{ \equal{#1}{} }
        {\textcolor{magenta}{(Tali) #2}}
        {\textcolor{magenta}{(Tali) \sout{#1}\xspace{}#2}}
        {\ignoretext{teal}{\sout{#1}\xspace{}#2}}
}

\newcommand{\omerb}[2][]{%
    \ifthenelse{ \equal{#1}{} }
        {\textcolor{blue}{(OmerB) #2}}
        {\textcolor{blue}{(OmerB) \sout{#1}\xspace{}#2}}
        {\ignoretext{teal}{\sout{#1}\xspace{}#2}}
}

\newcommand{\ow}[2][]{%
    \ifthenelse{ \equal{#1}{} }
        {\textcolor{orange}{(Oliver) #2}}
        {\textcolor{orange}{(Oliver) \sout{#1}\xspace{}#2}}
        {\ignoretext{teal}{\sout{#1}\xspace{}#2}}
}

\newcommand{\tomer}[2][]{%
    \ifthenelse{ \equal{#1}{} }
        {\textcolor{brown}{(Tomer) #2}}
        {\textcolor{brown}{(Tomer) \sout{#1}\xspace{}#2}}
        {\ignoretext{teal}{\sout{#1}\xspace{}#2}}
}

\newcommand{\ariel}[2][]{%
    \ifthenelse{ \equal{#1}{} }
        {\textcolor{pink}{(Ariel) #2}}
        {\textcolor{pink}{(Ariel) \sout{#1}\xspace{}#2}}
        {\ignoretext{teal}{\sout{#1}\xspace{}#2}}
}

\newcommand{\cih}[2][]{%
    \ifthenelse{ \equal{#1}{} }
        {\textcolor{green}{(Charles) #2}}
        {\textcolor{green}{(Charles) \sout{#1}\xspace{}#2}}
}

\newcommand{\junhwa}[2][]{%
    \ifthenelse{ \equal{#1}{} }
        {\textcolor{violet}{(Junhwa) #2}}
        {\textcolor{violet}{(Junhwa) \sout{#1}\xspace{}#2}}
}

\newcommand{\shiranz}[2][]{%
    \ifthenelse{ \equal{#1}{} }
        {\textcolor{blue}{(Shiran) #2}}
        {\textcolor{blue}{(Shiran) \sout{#1}\xspace{}#2}}
}

\newcommand{\yuanzhen}[2][]{%
    \ifthenelse{ \equal{#1}{} }
        {\textcolor{cyan}{(Yuanzhen) #2}}
        {\textcolor{cyan}{(Yuanzhen) \sout{#1}\xspace{}#2}}
}

\newcommand{\miki}[2][]{%
    \ifthenelse{ \equal{#1}{} }
        {\textcolor{red}{(Miki) #2}}
        {\textcolor{red}{(Miki) \sout{#1}\xspace{}#2}}
}

\newcommand{\inbar}[2][]{%
    \ifthenelse{ \equal{#1}{} }
        {\textcolor{red}{(inbar) #2}}
        {\textcolor{red}{(inbar) \sout{#1}\xspace{}#2}}
}

    \newcommand{\ds}[1]{{\leavevmode\color[rgb]{0.8,0,0.8}[Deqing: #1]}}
\else
 \newcommand{\hila}[2][1]{}

\newcommand{\tali}[1]{}

\newcommand{\omerb}[1]{}

\newcommand{\ow}[1]{}

\newcommand{\tomer}[1]{}

\newcommand{\ariel}[1]{}

\newcommand{\cih}[1]{}

\newcommand{\junhwa}[1]{}

\newcommand{\yuanzhen}[1]{}

\newcommand{\miki}[1]{}
    \newcommand{\ds}[1]{}
\fi

\newcommand{\aftertabcaption}{\vskip 0.15in}  
\newcommand{\aftertab}{\vskip -0.1in}  

\usepackage[accepted]{icml2023}

\usepackage{amsmath}
\usepackage{amssymb}
\usepackage{mathtools}
\usepackage{amsthm}
\usepackage{caption}

\usepackage[capitalize,noabbrev]{cleveref}
\Crefname{figure}{Fig.}{Fig.}
\Crefname{section}{Sec.}{Sec.}

\theoremstyle{plain}

\theoremstyle{definition}

\theoremstyle{remark}

\usepackage[textsize=tiny]{todonotes}

\newcommand{\methodname}{Lumiere}

\icmltitlerunning{Lumiere: A Space-Time Diffusion Model for Video Generation}

\begin{document}

\twocolumn[
\icmltitle{Lumiere: A Space-Time Diffusion Model for Video Generation}



\icmlsetsymbol{equal}{*}
\icmlsetsymbol{equal_technical}{$\dagger$}

\begin{icmlauthorlist}
\icmlauthor{Omer Bar-Tal}{equal,google,weizmann}
\icmlauthor{Hila Chefer}{equal,google,tau}
\icmlauthor{Omer Tov}{equal,google}
\icmlauthor{Charles Herrmann}{equal_technical,google}
\icmlauthor{Roni Paiss}{equal_technical,google}
\icmlauthor{Shiran Zada}{equal_technical,google}
\icmlauthor{Ariel Ephrat}{equal_technical,google}
\icmlauthor{Junhwa Hur}{equal_technical,google}
\icmlauthor{Guanghui Liu}{google}
\icmlauthor{Amit Raj}{google}
\icmlauthor{Yuanzhen Li}{google}
\icmlauthor{Michael Rubinstein}{google}
\icmlauthor{Tomer Michaeli}{google,technion}
\icmlauthor{Oliver Wang}{google}
\icmlauthor{Deqing Sun}{google}
\icmlauthor{Tali Dekel}{google,weizmann}
\icmlauthor{Inbar Mosseri}{equal_technical,google}\\
\vspace{1em}
\textsuperscript{\rm 1}Google Research \quad \textsuperscript{\rm 2}Weizmann Institute \quad \textsuperscript{\rm 3}Tel-Aviv University \quad \textsuperscript{\rm 4}Technion \vspace{0.2cm}

\end{icmlauthorlist}


\icmlcorrespondingauthor{Firstname1 Lastname1}{first1.last1@xxx.edu}
\icmlcorrespondingauthor{Firstname2 Lastname2}{first2.last2@www.uk}

\icmlkeywords{Machine Learning, ICML}

\vskip 0.1in

{

\begin{center}
    \centering
    \captionsetup{type=figure}
    \includegraphics[width=0.98\textwidth]
    {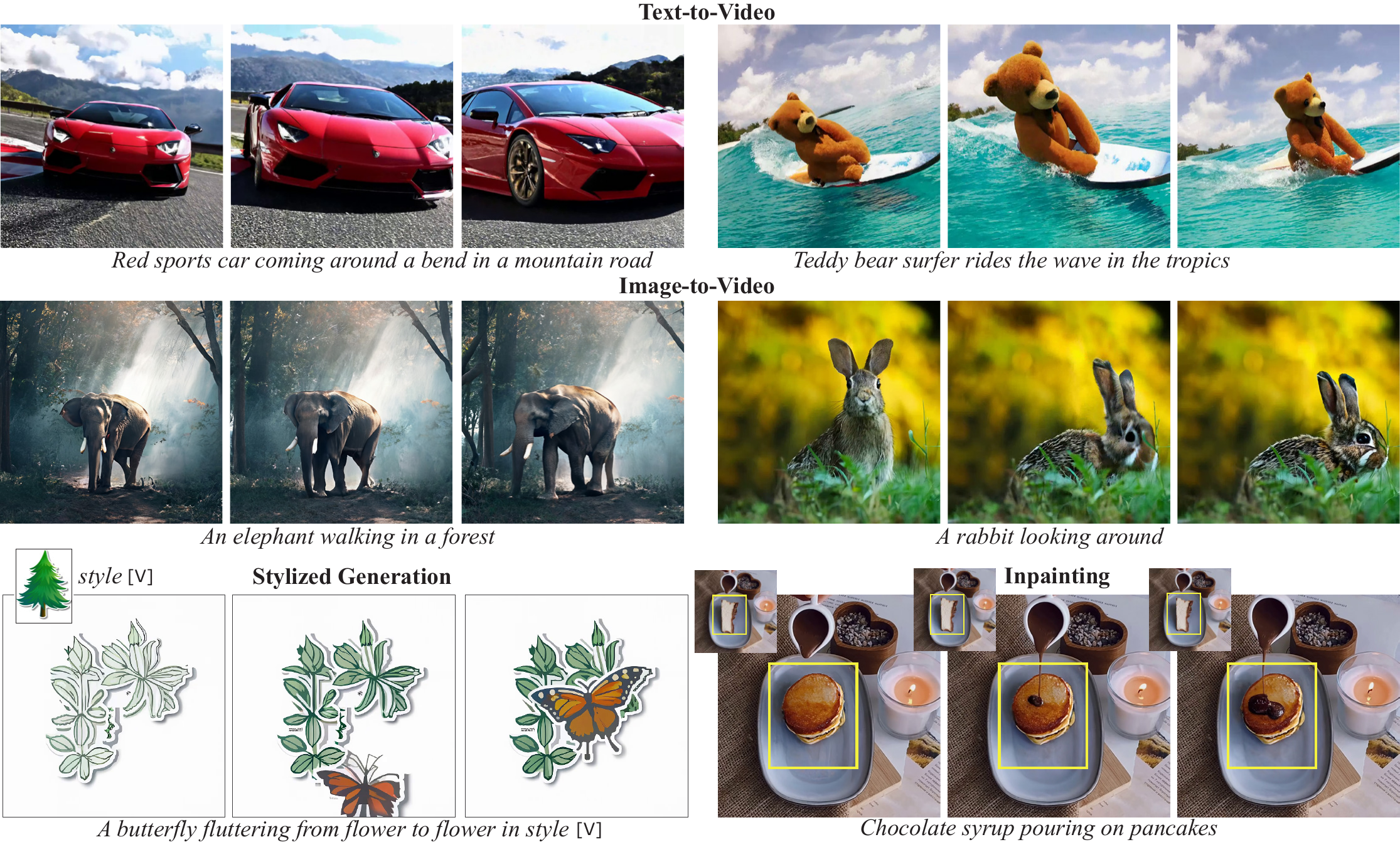}
    \captionof{figure}{\protect{Sample results generated by  \emph{Lumiere}, including text-to-video generation (first row),  image-to-video (second row), style-referenced generation, and video inpainting (third row; the bounding box indicates the inpainting mask region).} }
    \label{fig:teaser}
\end{center}%
\vskip 0.1in

}]





\let\thefootnote\relax\footnotetext{
\hspace*{-\footnotesep}
\textsuperscript{*}Equal first author
\textsuperscript{$\dagger$}Core technical contribution 
}
\relax\footnotetext{
\\
Work was done while O. Bar-Tal, H. Chefer were interns at Google.
Webpage: \url{https://lumiere-video.github.io/}
}

\begin{abstract}
We introduce \emph{\methodname{}} -- a text-to-video diffusion model designed for synthesizing videos that portray realistic, diverse and coherent motion -- a pivotal challenge in video synthesis. To this end, we introduce a Space-Time \mbox{U-Net} architecture that generates the entire temporal duration of the video \emph{at once}, through a single pass in the model. This is in contrast to existing video models which synthesize distant keyframes followed by temporal super-resolution -- an approach that inherently makes global temporal consistency difficult to achieve. By deploying both spatial and (importantly) temporal down- and up-sampling and leveraging a pre-trained text-to-image diffusion model, our model learns to directly generate a full-frame-rate, low-resolution video by processing it in multiple space-time scales. 
  We demonstrate state-of-the-art text-to-video generation results
, and show that our design easily facilitates a wide range of content creation tasks and video editing applications, including image-to-video, video inpainting, and stylized generation.


\begin{figure*}[t!h!]
    \centering
    \includegraphics[width=\textwidth]{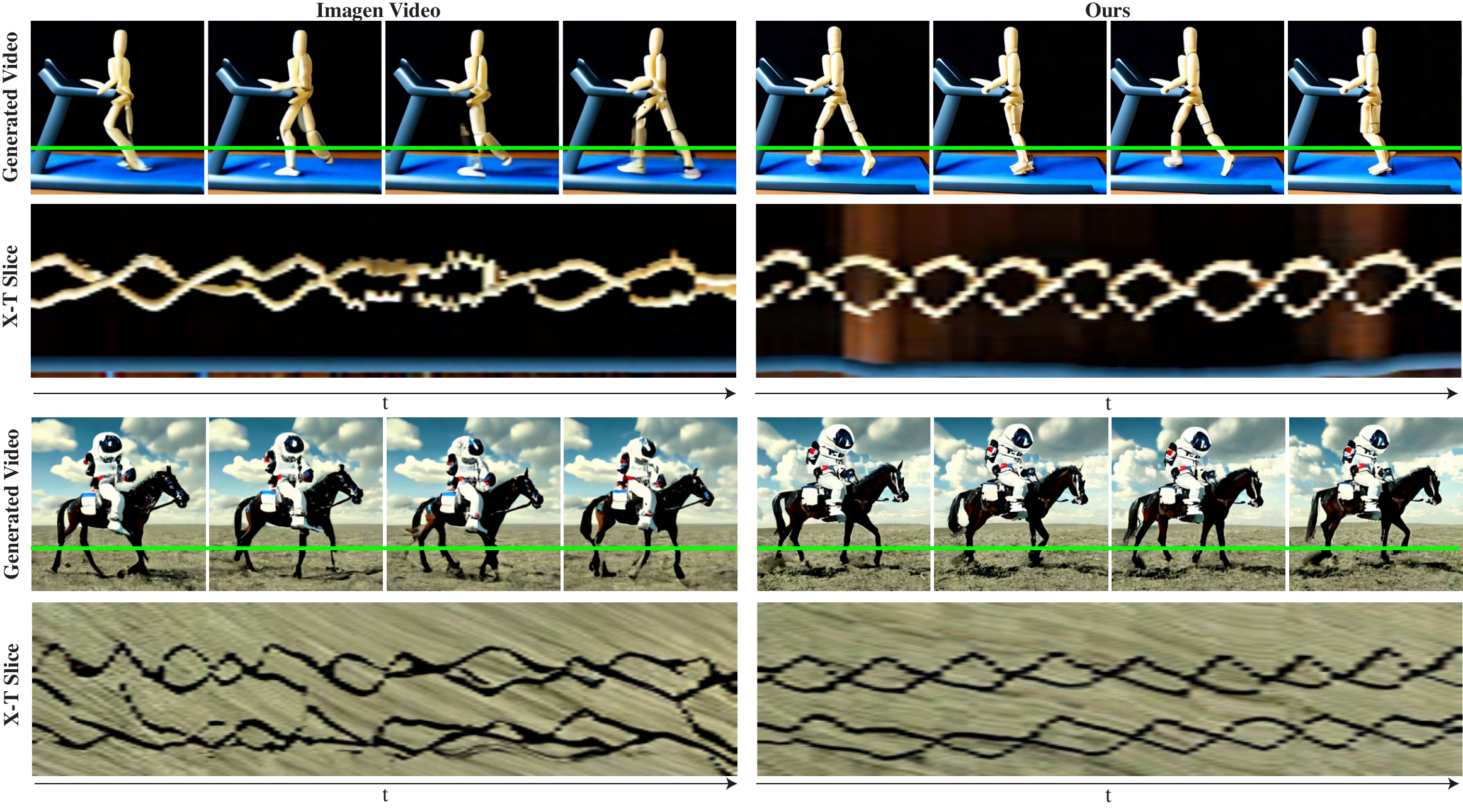}
        \caption{\textbf{Temporal consistency in generated videos.} Representative examples of generated videos using our model and ImagenVideo~\cite{ho2022imagen} for periodic motion. We apply Lumiere image-to-video generation, conditioned on the first frame of a video generated by ImagenVideo, and visualize corresponding X-T slices. ImagenVideo struggles to generate globally coherent repetitive motion due to its cascaded design and temporal super resolution modules, which fail to resolve aliasing ambiguities consistently across temporal windows. 
    }
    \label{fig:xt_slice}
\end{figure*}

\end{abstract}

\section{Introduction}
\label{sec:introduction}

\begin{figure*}[t!]
    \centering
    \includegraphics[width=\textwidth]{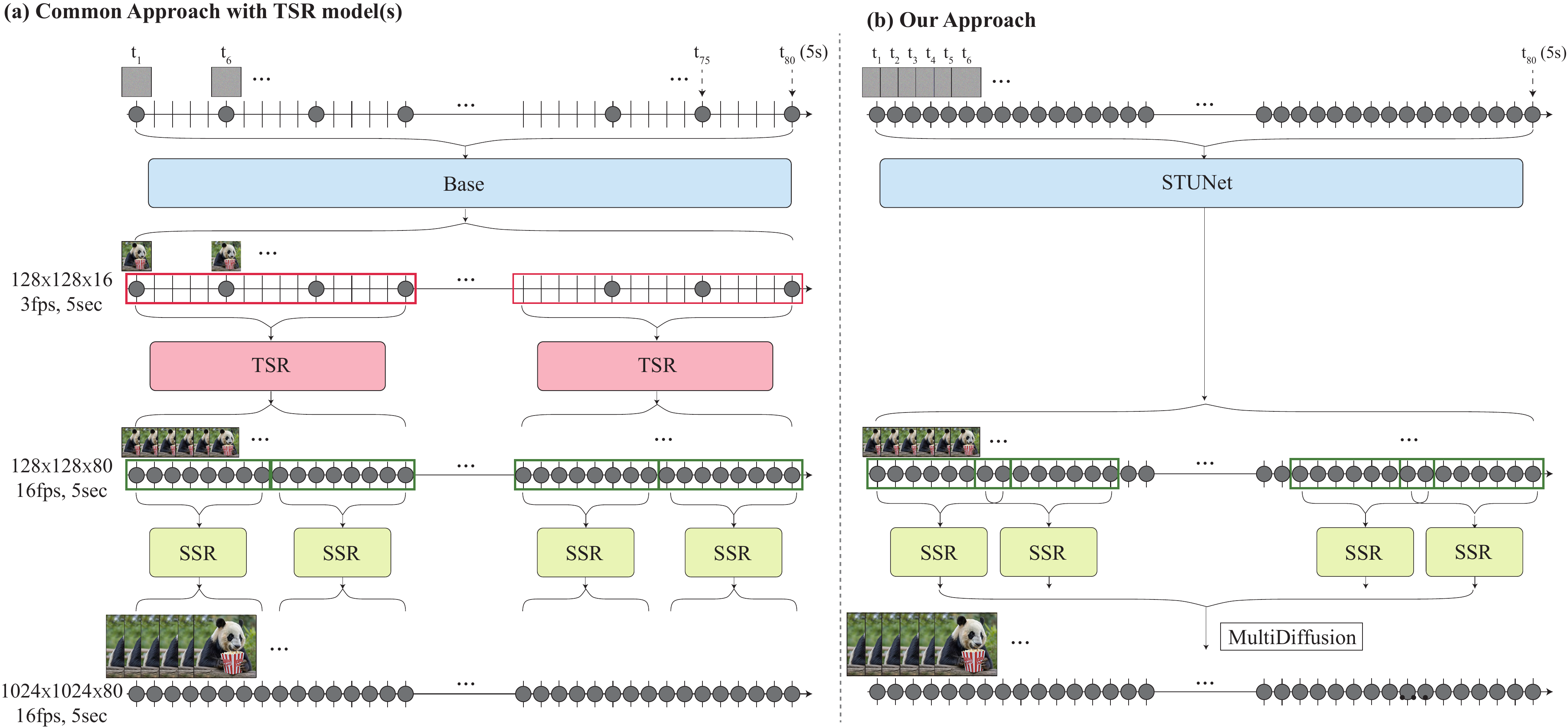}
    \caption{\textbf{Lumiere pipeline.} We illustrate our pipeline and the main difference from most common approach taken by previous works. (a) The common approach consists of a base model that generates distant keyframes, and a cascade of temporal super-resolution (TSR) models which subsequently fill in frames. A spatial super-resolution (SSR) model is applied in non-overlapping windows to obtain the high-resolution result. (b) In contrast, the base model in our framework process all frames \emph{at once}, without a cascade of TSR models, allowing us to learn globally coherent motion. To obtain the high-resolution video, we apply a SSR model on \emph{overlapping} windows and utilize MultiDiffusion \cite{bar2023multidiffusion} to combine the predictions into a coherent result. See Sec.\ref{sec:method} for details.
    }
    \label{fig:cascade}
\end{figure*}
Generative models for images have seen tremendous progress in recent years. State-of-the-art text-to-image (T2I) diffusion models are now capable of synthesizing high-resolution photo-realistic images that adhere to complex text prompts~\cite{saharia2022photorealistic,ramesh2022hierarchical,rombach2022high}, and allow a wide range of image editing capabilities \cite{po2023state} and other downstream uses.
However, training large-scale text-to-\emph{video} (T2V) foundation models remains an open challenge due to the added complexities that motion introduces. 
Not only are we sensitive to errors in modeling natural motion, but the added temporal data dimension introduces significant challenges in terms of memory and compute requirements, as well as the scale of the required training data to learn this more complex distribution. 
As a result, while T2V models are rapidly improving, existing models are still restricted in terms of video duration, overall visual quality, and the degree of realistic motion that they can generate.

A prevalent approach among existing T2V models is to adopt a cascaded design in which a base model generates distant keyframes, and subsequent temporal super-resolution (TSR) models generate the missing data between the keyframes in non-overlapping segments.
While memory efficient, the ability to generate globally coherent motion using temporal cascades is inherently restricted for the following reasons: (i) The base model generates an aggressively sub-sampled set of keyframes, in which fast motion becomes temporally aliased and thus ambiguous. (ii) TSR modules are constrained to fixed, small temporal context windows, and thus cannot consistently resolve aliasing ambiguities across the full duration of the video (illustrated in \cref{fig:xt_slice} in the case of synthesizing  periodic motion, e.g., walking). (iii) Cascaded training regimens in general suffer from a domain gap, where the TSR model is trained on real downsampled video frames, but at inference time is used to interpolate generated frames, which accumulates errors. 

Here, we take a different approach by introducing a new T2V diffusion framework that generates the full temporal duration of the video \emph{at once}.  
We achieve this by using a Space-Time U-Net (STUNet) architecture that learns to downsample the signal in both space \emph{and time}, and performs the majority of its computation in a compact space-time representation. 
This approach allows us to generate 80 frames at 16fps (or 5 seconds, which is longer than the average shot duration in most media~\cite{cutting2015shot}) with a single base model, leading to more globally coherent motion compared to prior work. 
Surprisingly, this design choice has been overlooked by previous T2V models, which follow the convention to include \emph{only spatial} down- and up-sampling operations in the architecture, and maintain a \emph{fixed temporal resolution} across the network \cite{ho2022video,ho2022imagen,singer2022make,ge2023preserve,blattmann2023videoldm,wang2023modelscope,guo2023animatediff,zhang2023show,girdhar2023emu,po2023state}. 

To benefit from the powerful generative prior of T2I models, we follow the trend of building \methodname{} on top of a pretrained (and fixed) T2I model~\cite{hong2022cogvideo,singer2022make,saharia2022photorealistic}.
In our case, the T2I model works in pixel space and consists of a base model followed by a \emph{spatial} super-resolution (SSR) cascade.
Since the SSR network operates at high spatial resolution, applying it on the entire video duration is infeasible in terms of memory requirements.
Common SSR solutions use a temporal windowing approach, which splits the video into non-overlapping segments and stitches together the results. However, this can lead to inconsistencies in appearance at the boundaries between windows~\cite{girdhar2023emu}.  
We propose to extend Multidiffusion~\cite{bar2023multidiffusion}, an approach proposed for achieving global continuity in panoramic image generation, to the temporal domain, where we compute spatial super-resolution on temporal windows, and aggregate results into a globally coherent solution over the whole video clip.


We demonstrate state-of-the-art video generation results and show how to easily adapt Luimere to a plethora of video content creation tasks, including video inpainting (\cref{fig:inpainting}), image-to-video generation (\cref{fig:results}), or generating stylized videos that comply with a given style image (\cref{fig:styledrop}).
Finally, we demonstrate that generating the full video at once allows us to easily invoke off-the-shelf editing methods to perform consistent editing (\cref{fig:sdedit}).

\begin{figure*}[t!]
    \centering
    \includegraphics[width=\textwidth]{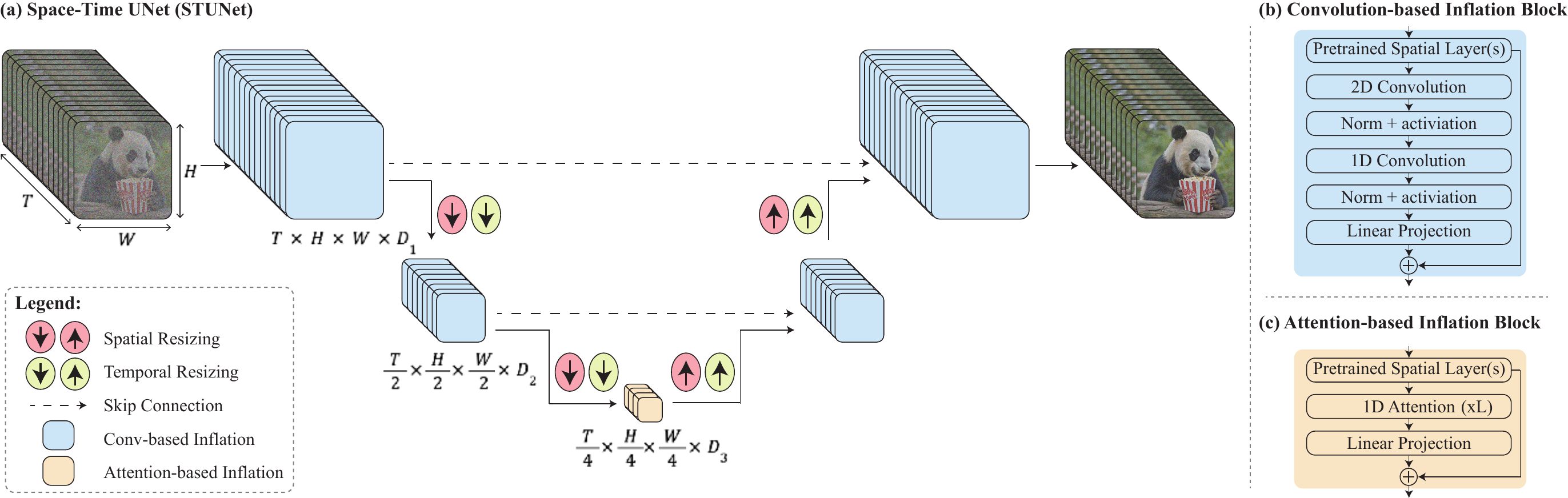}
    \caption{\textbf{STUNet architecture.}  We ``inflate'' a pre-trained T2I U-Net architecture \cite{ho2022imagen} into a Space-Time UNet (STUNet) that down- and up-sample the video in both space and time. (a) An illustration of STUNet's activation maps; color indicates features resulting from different temporal modules: (b) \emph{Convolution-based} blocks which consist of pre-trained T2I layers followed by a factorized space-time convolution, and (c) \emph{Attention-based} blocks at the coarsest U-Net level in which the pre-trained T2I layers are followed by temporal attention. Since the video representation is compressed at the coarsest level, we stack several temporal attention layers with limited computational overhead. See \cref{sec:stunet} for  details.}
    \label{fig:architecture}
\end{figure*}

\section{Related work}
\label{sec:related}

\paragraph{Text-to-Image Generation.}
Most of the common approaches for text-to-image (T2I) generation are based on diffusion models~\cite{sohl2015deep,ho2020denoising,song2020score}. 
Of these, DALL-E2~\cite{ramesh2022hierarchical} and Imagen \cite{saharia2022photorealistic} achieve photorealistic text-to-image generation using cascaded diffusion models, whereas Stable Diffusion ~\cite{rombach2022high} performs generation in a compressed low-dimensional latent space. 
A promising line of works design T2I diffusion models that generate high-resolution images end-to-end, without a spatial super-resolution cascaded system or fixed pre-trained latent space \cite{hoogeboom2023simple,gu2023matryoshka,chen2023importance}.
Here, we design a T2V model that generates the full frame duration at once, avoiding the \emph{temporal cascade} commonly involved in T2V models. 
\paragraph{Text-to-Video Generation.}
Recently, there have been substantial efforts in training large-scale T2V models on large scale datasets with autoregressive Transformers (e.g., \cite{villegas2022phenaki,wu2022nuwa,hong2022cogvideo,kondratyuk2023videopoet}) or Diffusion Models (e.g., \cite{ho2022imagen,ho2022video,gupta2023photorealistic}). A prominent approach for T2V generation is to ``inflate'' a pre-trained T2I model by inserting temporal layers to its architecture, and fine-tuning only those, or optionally the whole model, on video data \cite{singer2022make,blattmann2023videoldm,girdhar2023emu,ge2023preserve,Yuan2024Efficient}. PYoCo~\cite{ge2023preserve} carefully design video noise prior and obtain better performance for fine-tuning a T2I model for video generation. VideoLDM~\cite{blattmann2023videoldm} and AnimateDiff~\cite{guo2023animatediff} inflate StableDiffusion \cite{rombach2022high} and train only the newly-added temporal layers, showing they can be combined with the weights of personalized T2I models.
Interestingly, the ubiquitous convention of existing inflation schemes is to maintain a \emph{fixed temporal resolution} across the network, which limits their ability to process full-length clips. In this work, we design a new inflation scheme which includes learning to downsample the video in both space \emph{and time}, and performing the majority of computation in the compressed space-time feature space of the network. 
We extend an Imagen T2I model \cite{saharia2022photorealistic}, however our architectural contributions could be used for latent diffusion as well, and are orthogonal to possible improvements to the diffusion noise scheduler \cite{ge2023preserve} or to the video data curation \cite{blattmann2023stable}. 

\section{Lumiere}
\label{sec:method}

We utilize Diffusion Probabilistic Models as our generative approach \cite{sohl2015deep,croitoru2022diffusion,beatgan,ho2020denoising,nichol2021improved}. These models are trained to approximate a data distribution (in our case, a distribution over videos) through a series of denoising steps. Starting from a Gaussian i.i.d.~noise sample, the diffusion model gradually denoises it until reaching a clean sample drawn from the approximated target distribution. Diffusion models can learn a conditional distribution by incorporating additional guiding signals, such as text embedding, or spatial conditioning (e.g., depth map) \cite{beatgan,saharia2022palette,croitoru2023diffusion,zhang2023adding}.

Our framework consists of a base model and a spatial super-resolution (SSR) model. As illustrated in Fig.~\ref{fig:cascade}b, our base model generates full clips at a coarse spatial resolution. 
The output of our base model is spatially upsampled using a temporally-aware SSR model, resulting with the high-resolution video. 
We next describe the key design choices in our architecture, and demonstrate the applicability of our framework for a variety of downstream applications.

\begin{figure*}
    \centering
    \includegraphics[width=\textwidth]{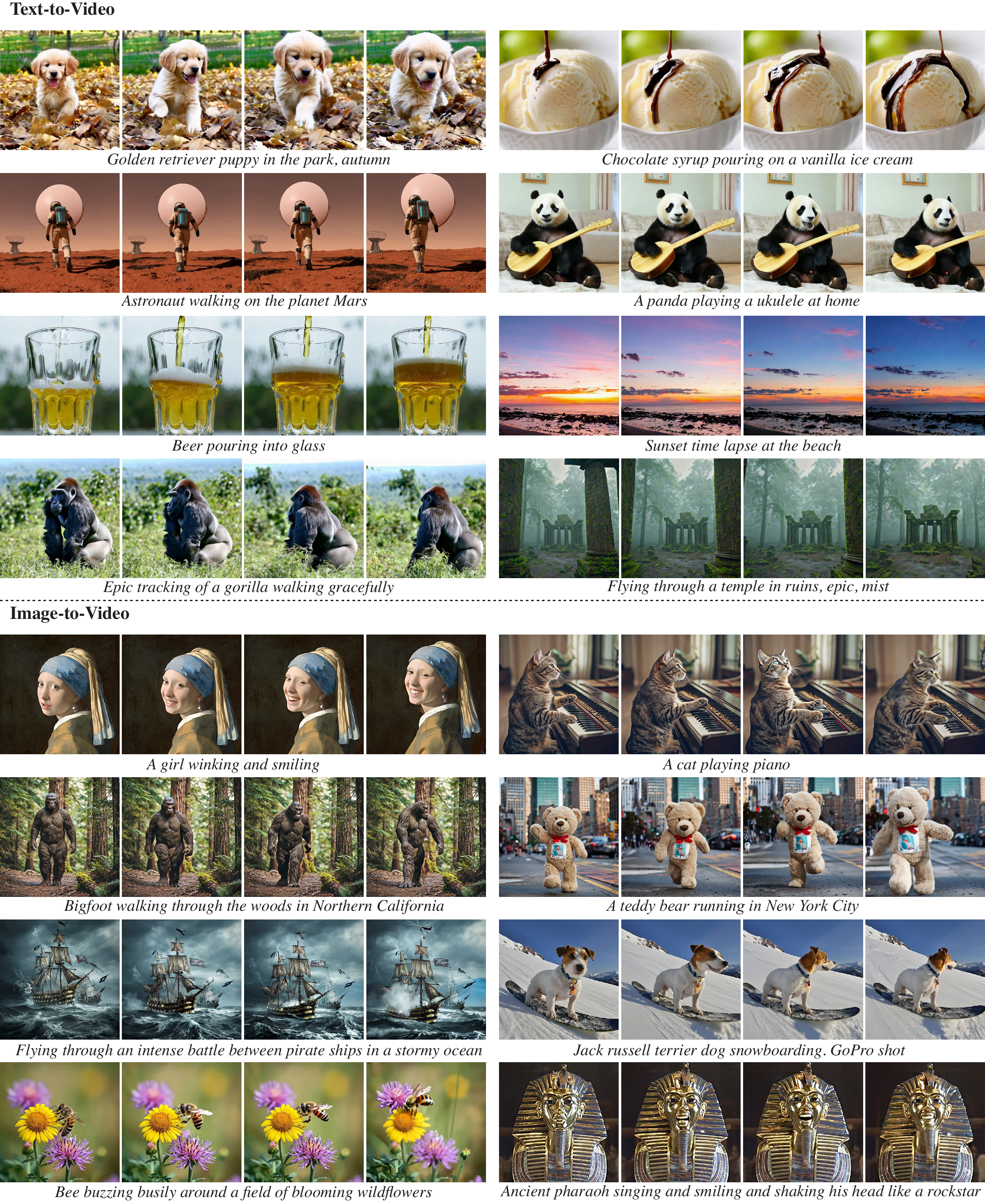}
    \caption{\textbf{Video generation results.} Sample results of text-to-video and image-to-video generation. The text prompt is indicated below each example. For image-to-video, the leftmost frame is provided to the model as a condition (see Sec.~\ref{sec:img2vid}). We refer the reader to the SM for full-video results.}
    \label{fig:results}
\end{figure*}

\subsection{Space-Time U-Net (STUnet)}
\label{sec:stunet}

To make our problem computationally tractable, we propose to use a space-time U-Net which downsamples the input signal both spatially \emph{and} temporally, and performs the majority of its computation on this compact space-time representation. 
We draw inspiration from \citet{3dunet}, who generalize the U-Net architecture \cite{ronneberger2015u} to include 3D pooling operations for efficient processing of volumetric biomedical data. 

Our architecture is illustrated in Fig.~\ref{fig:architecture}. We interleave temporal blocks in the T2I architecture, and insert temporal down- and up-sampling modules following each pre-trained spatial resizing module (Fig.~\ref{fig:architecture}a).
The temporal blocks include temporal convolutions (Fig.~\ref{fig:architecture}b) and temporal attention (Fig.~\ref{fig:architecture}c). Specifically, in all levels except for the coarsest, we insert factorized space-time convolutions (Fig.~\ref{fig:architecture}b) which allow increasing the non-linearities in the network compared to full-3D convolutions while reducing the computational costs, and increasing the expressiveness compared to 1D convolutions~\cite{tran2018closer}. As the computational requirements of temporal attention scale quadratically with the number of frames, we incorporate temporal attention only at the coarsest resolution, which contains a space-time compressed representation of the video.
Operating on the low dimensional feature map allows us to stack several temporal attention blocks with limited computational overhead. 

Similarly to \cite{blattmann2023videoldm,guo2023animatediff}, we train the newly added parameters, and keep the weights of the pre-trained T2I fixed. Notably, the common inflation approach ensures that at initialization, the T2V model is equivalent to the pre-trained T2I model, i.e., generates videos as a collection of independent image samples. However, in our case, it is impossible to satisfy this property due to the temporal down- and up-sampling modules. We empirically found that initializing these modules such that they perform nearest-neighbor down- and up- sampling operations results with a good starting point (see App.~\ref{app:init}).


\subsection{Multidiffusion for Spatial-Super Resolution}
\label{sec:multidiffusion}
Due to memory constraints, the inflated SSR network can operate only on short segments of the video.
To avoid temporal boundary artifacts, we achieve smooth transitions between the temporal segments by employing Multidiffusion~\cite{bar2023multidiffusion} along the temporal axis. At each generation step, we split the noisy input video $\smash{J\in \mathbb R^{H\times W \times T \times 3}}$ into a set of overlapping segments $\smash{\{J_i\}_{i=1}^N}$, where $\smash{J_i\in \mathbb R^{H\times W \times T' \times 3}}$ is the $i^{\text{th}}$ segment, which has temporal duration $T' < T$. To reconcile the per-segment SSR predictions, $\smash{\{\Phi(J_i)\}_{i=1}^N}$, we define the result of the denoising step to be the solution of the optimization problem
$$\argmin_{J'} \sum_{i=1}^n \left\| J' - \Phi(J_i) \right\|^2.$$
The solution to this problem is given by linearly combining the predictions over overlapping windows. See App.~\ref{app:ssr}.

\section{Applications} 
\label{sec:applications}

\begin{figure*}
    \centering
    \includegraphics[width=\textwidth]{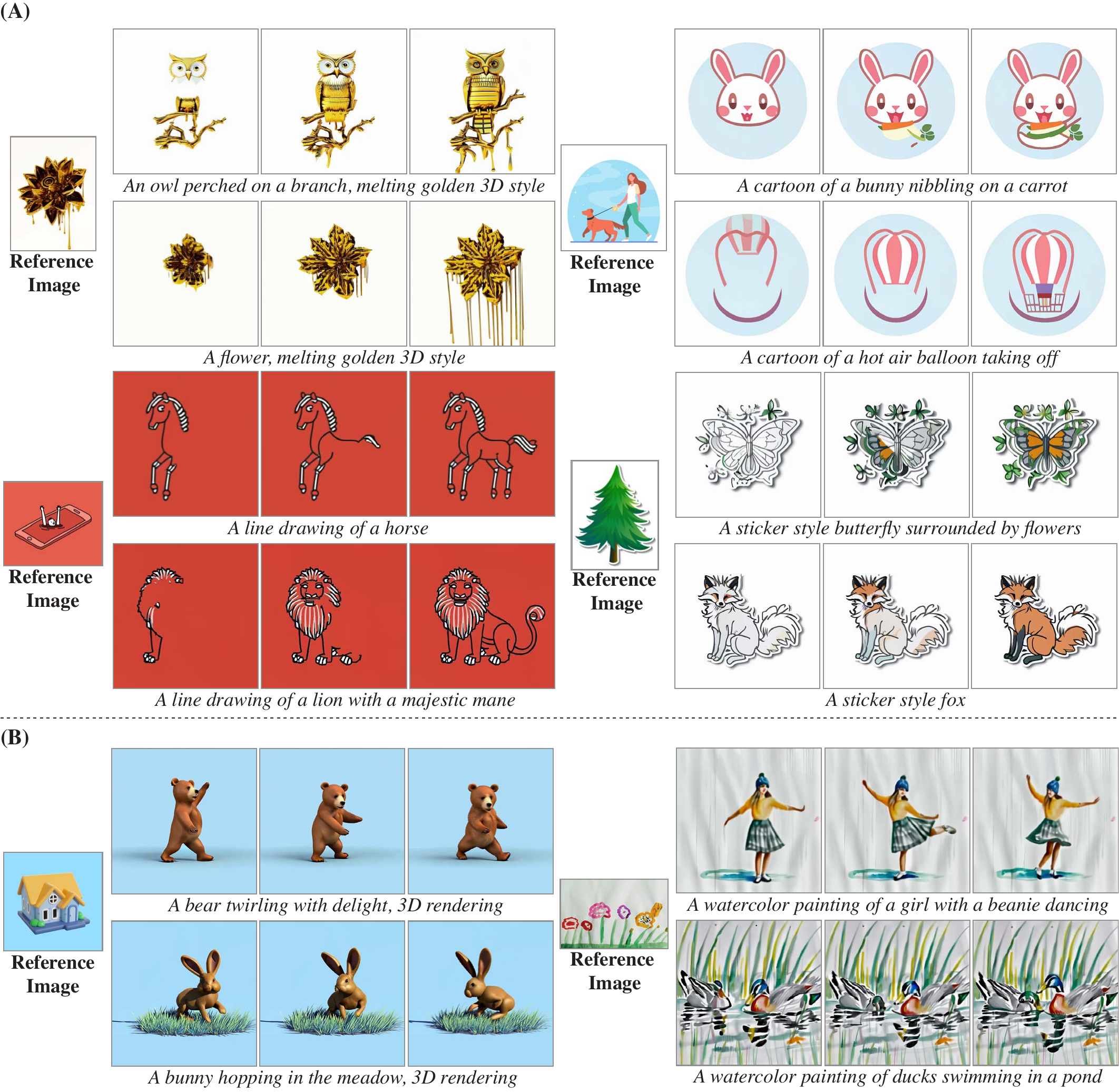}
    \caption{\textbf{Stylized Generation.} Given a driving style image and its corresponding set of fine-tuned text-to-image weights, we perform linear interpolation between the fine-tuned and pre-trained weights of the model's spatial layers. We present results for (A) vector art styles, and (B) realistic styles. The results demonstrate Lumiere's ability to creatively match a \emph{different} motion prior to each of the spatial styles (frames shown from left to right). See Sec.~\ref{sec:styledrop} for details.
    }
    \label{fig:styledrop}
\end{figure*}

\begin{figure*}
    \centering
    \includegraphics[width=\textwidth]{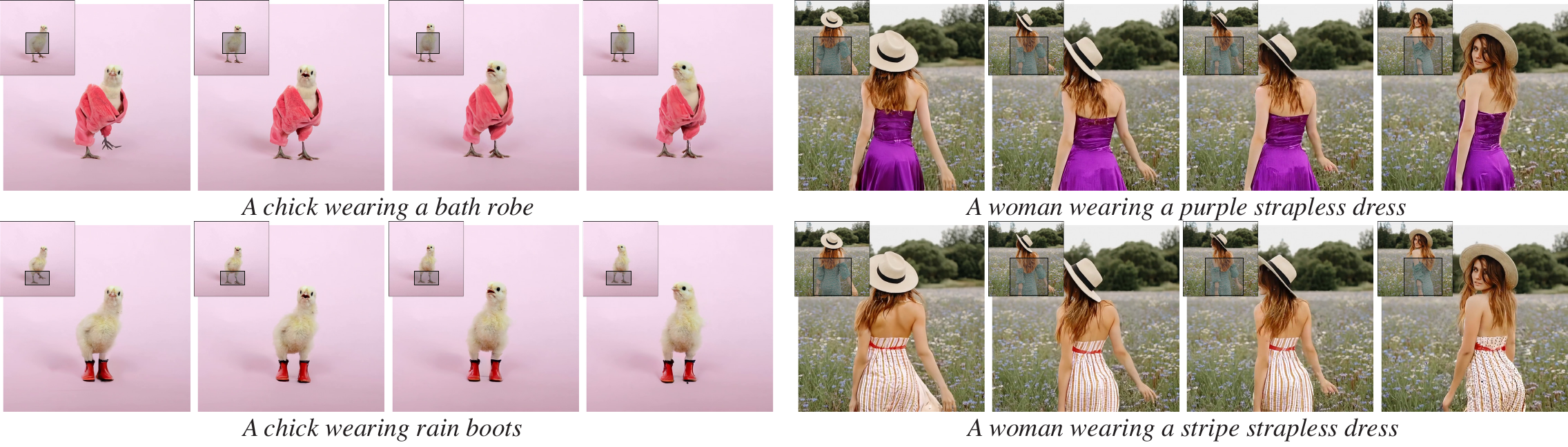}
    \caption{\textbf{Inpainting.} Examples of video inpainting with Lumiere. For each input video (top left corner of each frame), we animate the masked area of the video using our model.}
    \label{fig:inpainting}
\end{figure*}

\begin{figure*}
    \centering
    \vspace{0.5em}
    \includegraphics[width=\textwidth]{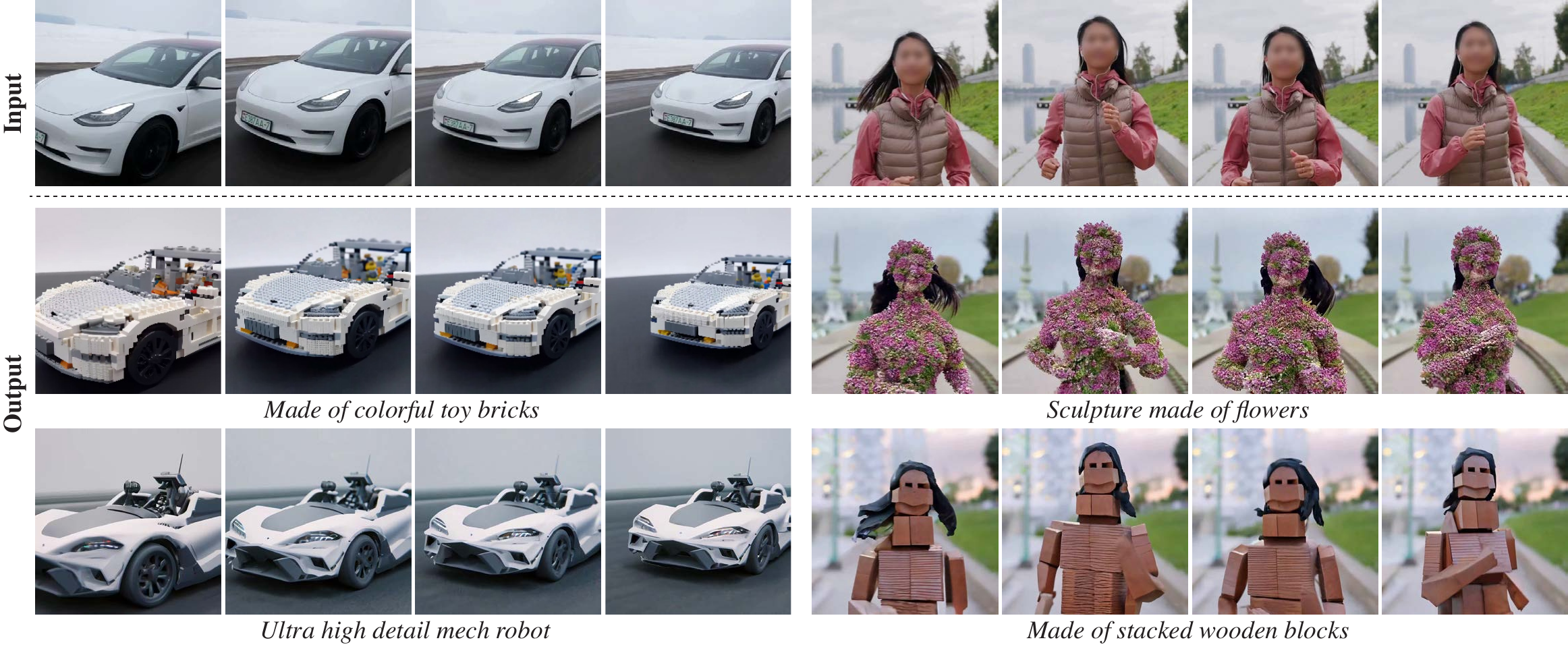}
    \vspace{0.5em}
    \caption{\textbf{Video-to-video via SDEdit.} Our base model generates full-frame-rate videos, without a TSR cascade, and thus facilitates an intuitive interface for downstream applications. We demonstrate this property by applying SDEdit \cite{meng2022sdedit} using our model, achieving consistent video stylization. We show several frames of a given input video in the first row, and the corresponding edited frames below.}
    \label{fig:sdedit}
\end{figure*}

The lack of a TSR cascade makes it easier to extend \methodname{} to downstream applications. In particular, our model provides an intuitive interface for downstream applications that require an off-the-shelf T2V model (e.g., \citet{meng2022sdedit,poole2022dreamfusion,gal2023breathing}). We demonstrate this property by performing video-to-video editing using SDEdit~\cite{meng2022sdedit} (see Fig.~\ref{fig:sdedit}).
We next discuss a number of such applications, including style conditioned generation, image-to-video, inpainting and outpainting, and cinemagraphs. 
We present example frames in Figs.~\ref{fig:styledrop}-\ref{fig:sdedit} and refer the reader to the Supplementary Material (SM) on our webpage for full video results.
\subsection{Stylized Generation}
\label{sec:styledrop}
Recall that we only train the newly-added temporal layers and keep the pre-trained T2I weights fixed.
Previous work showed that substituting the T2I weights with a  model customized for a specific style allows to generate videos with the desired style \cite{guo2023animatediff}.
We observe that this simple ``plug-and-play'' approach often results in distorted or static videos (see SM), and hypothesize that this is caused by the significant deviation in the distribution of the input to the temporal layers from the fine-tuned spatial layers. 

Inspired by the success of GAN-based interpolation approaches~\cite{Pinkney2020ResolutionDG}, we opt to strike a balance between style and motion by linearly interpolating between the fine-tuned T2I weights, $W_{\text{style}}$, and the original T2I weights, $W_{\text{orig}}$. Specifically, we construct the interpolated weights as $W_{\text{interpolate}} = \alpha \cdot W_{\text{style}} + (1-\alpha) \cdot W_{\text{orig}}$. The interpolation coefficient $\alpha \in [0.5,1]$ is chosen manually in our experiments to generate videos that  adhere to the style and depict plausible motion. 

Figure~\ref{fig:styledrop} presents sample results for various styles from \cite{sohn2023styledrop}. 
While more realistic styles such as ``watercolor painting'' result in realistic motion, other, less realistic spatial priors derived from vector art styles, result in corresponding unique non-realistic motion. For example, the ``line drawing'' style results in animations that resemble pencil strokes ``drawing'' the described scene, while the ``cartoon'' style results in content that gradually  ``pops out'' and constructs the scene (see SM for full videos).

\subsection{Conditional Generation}
Similarly to \citet{blattmann2023videoldm,wang2023videocomposer}, we extend our model to video generation conditioned on additional input signals (e.g., image or mask). 
We achieve this by modifying the model to take as input two signals in addition to the noisy video $\smash{J\in \mathbb{R}^{T \times H \times W \times 3}}$ and the driving text prompt. Specifically, we add the masked conditioning video $\smash{C\in \mathbb{R}^{T \times H \times W \times 3}}$ and its corresponding binary mask $\smash{M\in \mathbb{R}^{T \times H \times W \times 1}}$, such that the overall input to the model is the concatenated tensor $\left<J,C,M\right> \in \mathbb{R}^{T \times H \times W \times 7}$. We expand the channel dimension of the first convolution layer from $3$ to $7$ in order to accommodate the modified input shape and fine-tune our base T2V model to denoise $J$ based on $C, M$. During this fine-tuning process, we take $J$ to be the noisy version of the training video, and $C$ to be a masked version of the clean video. This encourages the model to learn to copy the unmasked information in $C$ to the output video while only animating the masked content, as desired.

\paragraph{Image-to-Video.}
\label{sec:img2vid} In this case, the first frame of the video is given as input. The conditioning signal $C$ contains this first frame followed by blank frames for the rest of the video. The corresponding mask $M$ contains ones (i.e., unmasked content) for the first frame and zeros (i.e., masked content) for the rest of the video. Figures~\ref{fig:teaser} and \ref{fig:results} show sample results of image-conditioned generation (see SM for more results). Our model generates videos that start with the desired first frame, and exhibit intricate coherent motion across the entire video duration. 

\begin{figure}
    \centering
    \includegraphics[width=0.5\textwidth]{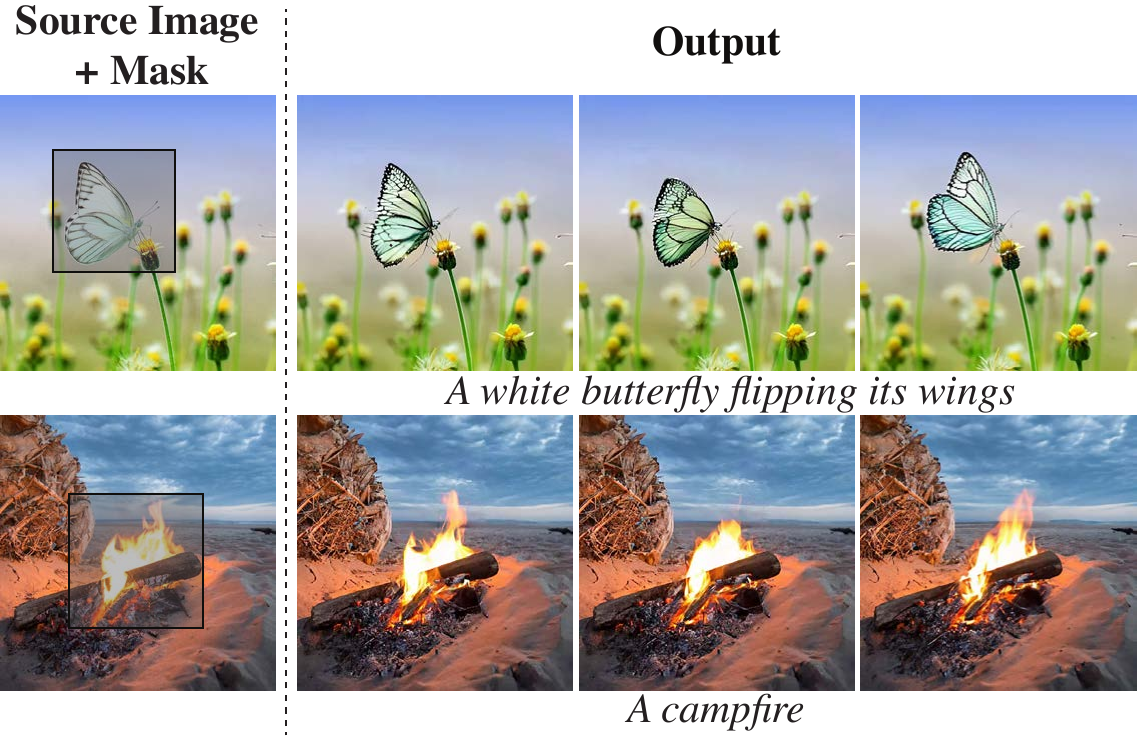}
    \caption{\textbf{Cinemagraphs.} Given only an input \emph{image} and a mask (left), our method generates a video in which the marked area is animated and the rest remains static (right).}
    \label{fig:motionbrush}
\end{figure}


\paragraph{Inpainting.} 
Here, the conditioning signals are a user-provided video $C$ and a mask $M$ that describes the region to complete in the video. Note that the inpainting application can be used for object replacement/insertion (Fig.~\ref{fig:teaser}) as well as for localized editing (Fig.~\ref{fig:inpainting}). The effect is a seamless and natural completion of the masked region, with contents guided by the text prompt. We refer the reader to the SM for more examples of both inpainting and outpainting.

\paragraph{Cinemagraphs.}
We additionally consider the application of animating the content of \emph{an image} only within a specific user-provided region. The conditioning signal $C$ is the input image duplicated across the entire video, while the mask $M$ contains ones for the entire first frame (i.e., the first frame is unmasked), and for the other frames, the mask contains ones only outside the user-provided region (i.e., the other frames are masked inside the region we wish to animate). We provide sample results in \cref{fig:motionbrush} and in the SM. Since the first frame remains unmasked, the animated content is encouraged to maintain the appearance from the conditioning image. 


\begin{table}[t]
\aftertabcaption
\centering
\scriptsize
\resizebox{0.98\linewidth}{!}{
\begin{tabularx}{0.8\columnwidth}{@{}X S[table-format=2.2] @{\hskip 1em} S[table-format=2.2] @{}}
\toprule
Method & {FVD $\downarrow$} & {IS $\uparrow$} \\ 
\midrule
MagicVideo \cite{zhou2022magicvideo} & 655.00 & {-} \\
Emu Video \cite{girdhar2023emu} & 606.20 & 42.70 \\
Video LDM \cite{blattmann2023videoldm} & 550.61 & 33.45 \\
Show-1 \cite{zhang2023show} & 394.46 & 35.42 \\
Make-A-Video \cite{singer2022make} & 367.23 & 33.00 \\
PYoCo \cite{ge2023preserve} & 355.19 & 47.76 \\
SVD \cite{blattmann2023stable} & 242.02 & {-} \\
\textbf{Lumiere (Ours)} & 332.49 & 37.54 \\
\bottomrule
\end{tabularx}}
\caption{Zero-shot text-to-video generation comparison on UCF101 \cite{soomro2012ucf101}. Our method achieves competitive FVD \cite{fvd} and IS \cite{IS} scores. See~\cref{sec:ucf}.}
\aftertab
\label{tab:quantitive_eval}
\end{table}


\section{Evaluation and Comparisons}
We train our T2V model on a dataset containing 30M videos along with their text caption. 
The videos are 80 frames long at 16 fps (5 seconds). The base model is trained at $128\times128$ and the SSR outputs $1024\times 1024$ frames. 
We evaluate our model on a collection of 109 text prompts describing diverse objects and scenes. The prompt list consists of 91 prompts used by prior works \cite{singer2022make,ho2022imagen,blattmann2023videoldm} and the rest were created by us (see App.~\ref{app:eval}). Additionally, we employ a zero-shot evaluation protocol on the UCF101 dataset \cite{soomro2012ucf101}, as detailed in Sec.~\ref{sec:quantitative}. 

We illustrate text-to-video generation in~Figs.~\ref{fig:teaser} and \ref{fig:results}. Our method generates high-quality videos depicting both intricate object motion (e.g., walking astronaut in Fig.~\ref{fig:results}) and coherent camera motion (e.g., car example in Fig.~\ref{fig:teaser}). We refer the reader to the SM for full-video results.
\begin{figure*}
    \centering
    \includegraphics[width=\textwidth]{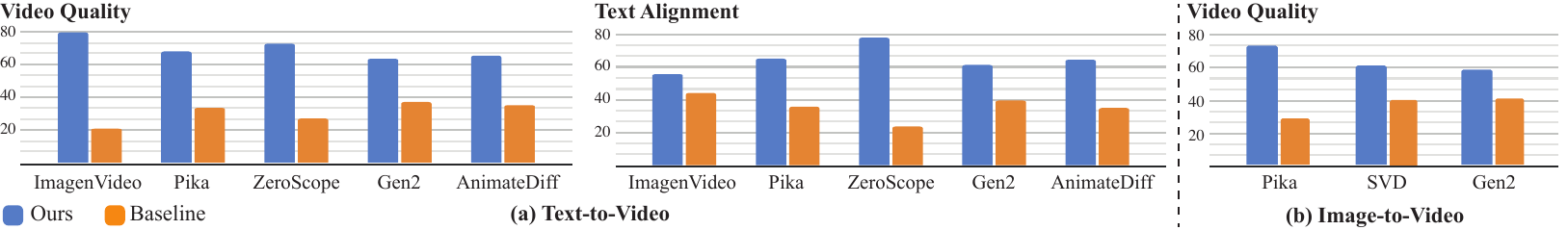}
    \caption{\textbf{User study.} We compare our method to each of the baselines. For each baseline, we report the percentage of user votes in our favor (blue) and in favor of the baseline (orange). Our method was preferred by users in both text-to-video, and image-to-video generation. See Sec.~\ref{sec:user_study}.}\label{fig:user_study}
\end{figure*}
\paragraph{Baselines.}
We compare our method to prominent T2V diffusion models: (i) ImagenVideo \cite{ho2022imagen}, that operates in pixel-space and consists of a cascade of 7 models (a base model, 3 TSR models, and 3 SSR models); (ii) AnimateDiff \cite{guo2023animatediff}, (iii) StableVideoDiffusion (SVD) \cite{blattmann2023stable}, and (iv) ZeroScope \cite{wang2023modelscope} that inflate Stable Diffusion \cite{rombach2022high} and train on video data; note that AnimateDiff and ZeroScope output only 16, and 36 frames respectively. SVD released only their image-to-video model, which outputs 25 frames and is not conditioned on text. Additionally, we compare to (v) Pika \cite{pika} and (vi) Gen-2 \cite{gen2} commercial T2V models that have available API.
Furthermore, we quantitatively compare to additional T2V models that are closed-source in Sec.~\ref{sec:quantitative}.
\label{sec:results}


\subsection{Qualitative Evaluation}\label{sec:qualitative_evaluation}
We provide qualitative comparison between our model and the baselines in \cref{fig:qualitative_comparison}. We observed that Gen-2 \cite{gen2} and Pika \cite{pika} demonstrate high per-frame visual quality; however, their outputs are characterized by a very limited amount of motion, often resulting in near-static videos. ImagenVideo \cite{ho2022imagen} produces a reasonable amount of motion, but at a lower overall visual quality. AnimateDiff \cite{guo2023animatediff} and ZeroScope \cite{wang2023modelscope} exhibit noticeable motion but are also prone to visual artifacts.
Moreover, they generate videos of shorter durations, specifically 2 seconds and 3.6 seconds, respectively.
In contrast, our method produces 5-second videos that have higher motion magnitude while maintaining temporal consistency and overall quality.

\subsection{Quantitative Evaluation}
\label{sec:quantitative}
\paragraph{Zero-shot evaluation on UCF101.}
\label{sec:ucf}
Following the evaluation protocols of \citet{blattmann2023stable} and \citet{ge2023preserve}, we quantitatively evaluate our method for zero-shot text-to-video generation on UCF101 \cite{soomro2012ucf101}. \cref{tab:quantitive_eval} reports the Fréchet Video Distance (FVD) \cite{fvd} and Inception Score (IS) \cite{IS} of our method and previous work. We achieve competitive FVD and IS scores. However, as discussed in previous work (e.g.,~\citet{girdhar2023emu,ho2022imagen,chong2020effectively}), these metrics do not faithfully reflect human perception, and may be significantly influenced by low-level details \cite{parmar2021cleanfid} and by the distribution shift between the reference UCF101 data and the T2V training data \cite{girdhar2023emu}. Furthermore, the protocol uses only 16 frames from generated videos and thus is not able to capture long-term motion.

\paragraph{User Study.}
\label{sec:user_study}
We adopt the Two-alternative Forced Choice (2AFC) protocol, as used in previous works \cite{kolkin2019style, zhang2018unreasonable, blattmann2023stable, rombach2022high}. In this protocol, participants were presented with a randomly selected pair of videos: one generated by our model and the other by one of the baseline methods. Participants were then asked to choose the video they deemed better in terms of visual quality and motion. Additionally, they were asked to select the video that more accurately matched the target text prompt. We collected $\sim$400 user judgments for each baseline and question, utilizing the Amazon Mechanical Turk (AMT) platform. As illustrated in \cref{fig:user_study}, our method was preferred over all baselines by the users and demonstrated better alignment with the text prompts. Note that ZeroScope and AnimateDiff generate videos only at 3.6 and 2 second respectively, we thus trim our videos to match their duration when comparing to them. 

We further conduct a user study for comparing our image-to-video model (see Sec.~\ref{sec:img2vid}) against Pika~\cite{pika}, StableVideoDiffusion (SVD)~\cite{blattmann2023stable}, and Gen2\cite{gen2}. Note that SVD image-to-video model is not conditioned on text, we thus focus our survey on the video quality. As seen in Fig.~\ref{fig:user_study}, our method was preferred by users compared to the baselines.
For a detailed description of the full evaluation protocol, please refer to~\cref{app:eval}.

\section{Conclusion}
\label{sec:conclusion}
\vspace{-0.3em}

We presented a new text-to-video generation framework, utilizing a pre-trained text-to-image diffusion model. We identified an inherent limitation in learning globally-coherent motion in the prevalent approach of first generating \emph{distant} keyframes and subsequently interpolating them using a cascade of temporal super-resolution models. To tackle this challenge, we introduced a space-time U-Net architecture design that directly generates full-frame-rate video clips, by incorporating both spatial, \emph{and temporal} down- and up-sampling modules. We demonstrated state-of-the-art generation results, and showed the applicability of our approach for a wide range of applications, including image-to-video, video inapainting, and stylized generation. 

As for limitations, our method is not designed to generate videos that consist of multiple shots, or that involve transitions between scenes. Generating such content remains an open challenge for future research. Furthermore, we established our model on top of a T2I model that operates in the pixel space, and thus involves a spatial super resolution module to produce high resolution images.  Nevertheless,  our design principles are applicable to latent video diffusion models \cite{rombach2022high}, and can trigger further research in the design of text-to-video models.
\vspace{-0.4em}
\section{Societal Impact}
\vspace{-0.4em}
Our primary goal in this work is to enable novice users to generate visual content in a creative and flexible way. However, there is a risk of misuse for creating fake or harmful content with our technology, and we believe that it is crucial to develop and apply tools for detecting biases and malicious use cases in order to ensure a safe and fair use.
\vspace{-0.5em}
\paragraph{Acknowledgments}
We would like to thank Ronny Votel, Orly Liba, Hamid Mohammadi, April Lehman, Bryan Seybold, David Ross, Dan Goldman, Hartwig Adam, Xuhui Jia, Xiuye Gu, Mehek Sharma, Rachel Hornung, Oran Lang, Jess Gallegos, William T. Freeman and David Salesin for their collaboration, helpful discussions, feedback and support.
We thank owners of images and videos used in our experiments for sharing their valuable assets (attributions can be found in our webpage).
\bibliography{short,example_paper}
\bibliographystyle{icml2023}

\newpage
\appendix
\onecolumn
\section*{Appendix}

\section{Qualitative Comparison}\label{app:qualitative_comparison}

We provide a qualitative comparison of our method and the baselines (see \cref{sec:qualitative_evaluation}). 

\begin{figure}[H]
    \centering
    \includegraphics[width=0.9\textwidth]{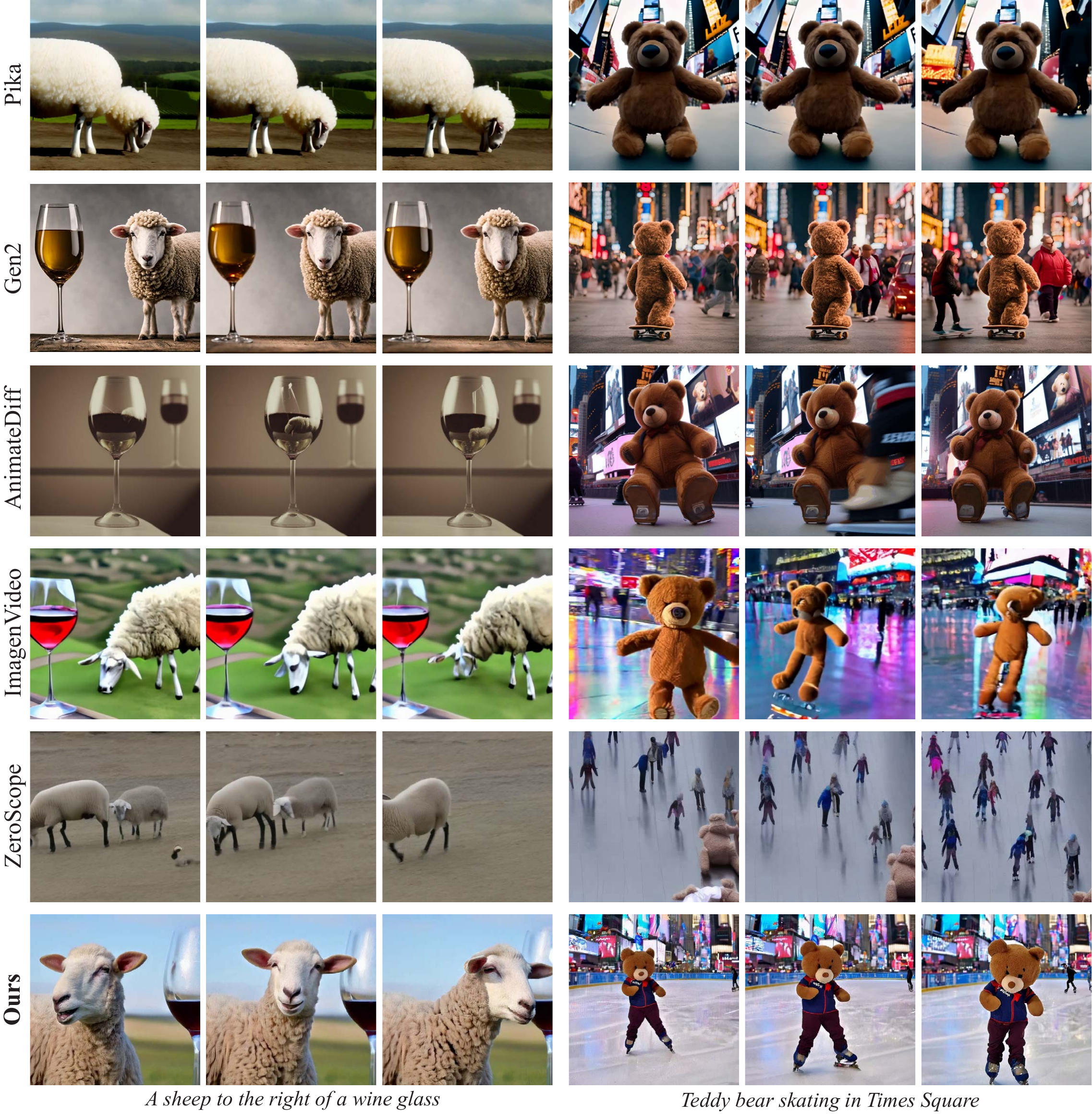}
    \caption{\textbf{Qualitative Comparison.} We provide samples from our user study, comparing our model to prominent T2V diffusion models, Gen-2 \cite{gen2}, Pika \cite{pika}, ImagenVideo \cite{ho2022imagen}, AnimateDiff \cite{guo2023animatediff} and ZeroScope \cite{wang2023modelscope}.}
    \label{fig:qualitative_comparison}
\end{figure}

\newpage
\section{Initialization}
\label{app:init}
\begin{figure}[H]
    \centering
    \includegraphics[width=0.6\textwidth]{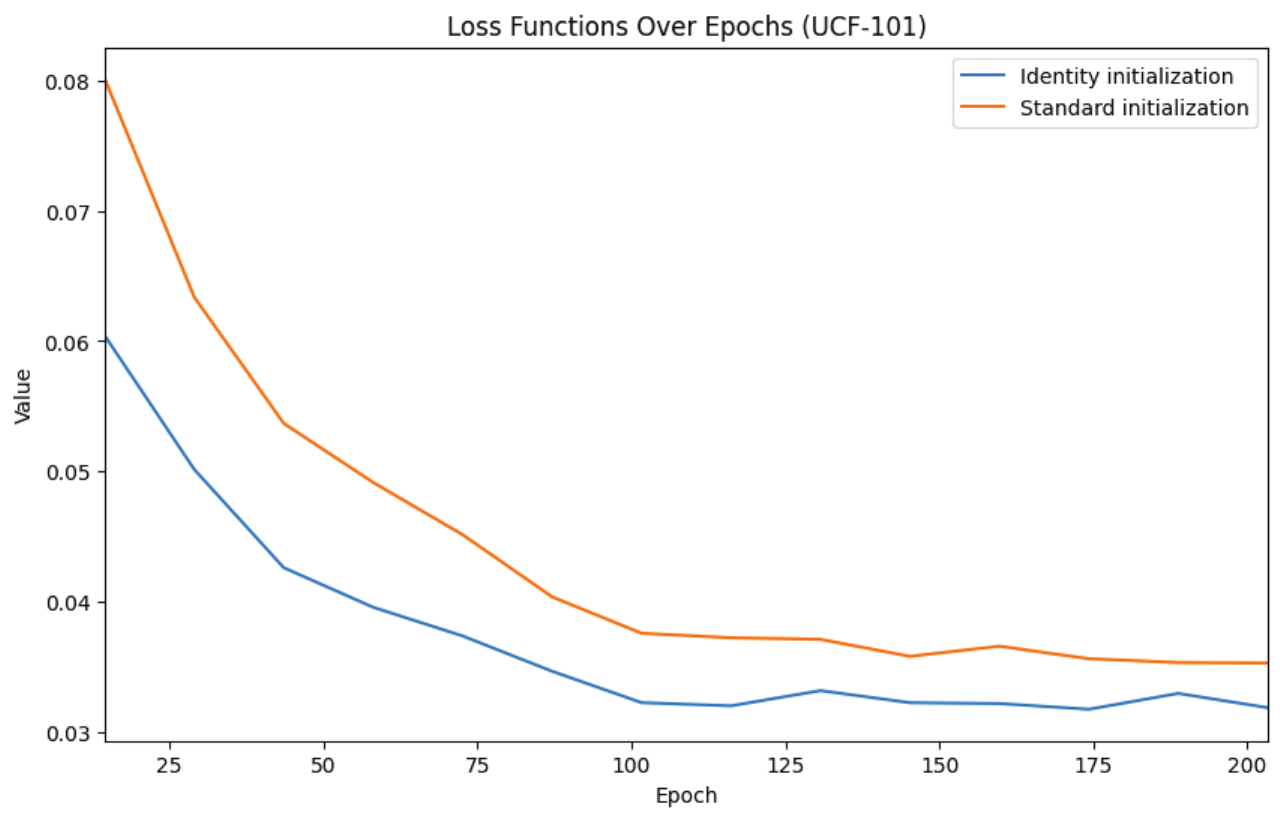}
    \caption{\textbf{Initialization Ablation.} We ablate the initialization of our temporal down- and up-sampling modules by training on UCF-101~\cite{soomro2012ucf101} and comparing the loss function. See App.~\ref{app:init} for details.}
    \label{fig:init}
\end{figure}

\begin{figure}[H]
    \centering
    \includegraphics[width=0.6\textwidth]{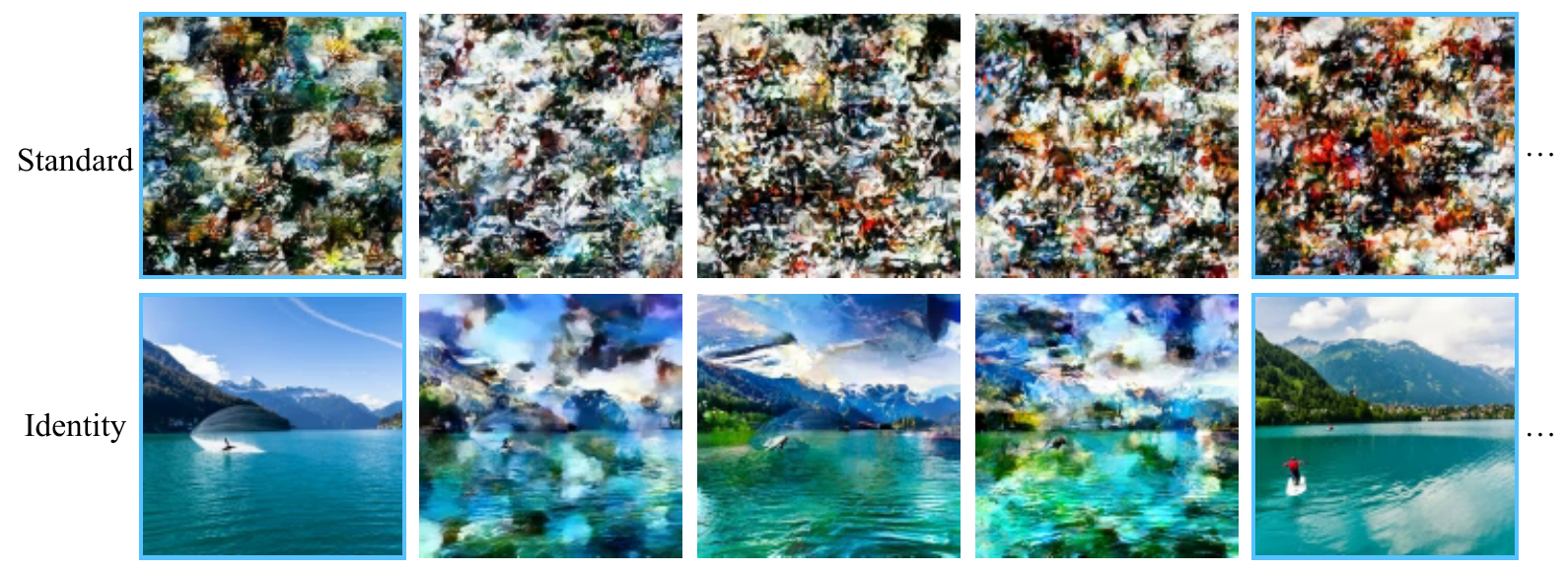}
    \caption{\textbf{Random Samples at Initialization.} We visualize several frames generated \emph{at initialization} with different initialization schemes for the temporal down- and up- sampling modules, to demonstrate their effect. The standard random initialization (first row) results with non-meaningful samples, because the temporal down- and up-sampling modules mix unrelated feature maps. The identity initialization (second row) allows us to better benefit from the prior of the pre-trained text-to-image. See App.~\ref{app:init} for details.}
    \label{fig:init_frames}
\end{figure}

As illustrated in Fig.~\ref{fig:architecture}(b)-(c), each inflation block has a residual structure, in which the temporal components are followed by a linear projection, and are added to the output of the pre-trained T2I spatial layer. Similarly to previous works (e.g., \cite{guo2023animatediff}), we initialize the linear projection to a zero mapping, such that at initialization, the inflation block is equivalent to the pre-trained T2I block. However, as mentioned in Sec.~\ref{sec:stunet}, due to the temporal down- and up- sampling modules, the overall inflated network cannot be equivalent at initialization to the pre-trained T2I model due to the temporal compression. We empirically found that initializing the temporal down- and up-sampling modules such that they perform nearest-neighbor down- and up- sampling operations (i.e., temporal striding or frame duplication) is better than a standard initialization~\cite{he2015delving}. In more detail, the temporal downsampling is implemented as a 1D temporal convolution (identity initialized) followed by a stride, and the temporal upsampling is implemented as frame duplication followed by a 1D temporal convolution (identity initialized). Note that this scheme ensures that at initialization, every $N^{\text{th}}$ frame is identical to the output of the pre-trained T2I model, where $N$ is the total temporal downsampling factor in the network (see Fig.~\ref{fig:init_frames}). We ablate our choice of temporal down- and up- sampling initialization on the UCF-101 \cite{soomro2012ucf101} dataset in Fig.~\ref{fig:init}.

\newpage
\section{SSR with MultiDiffusion}
\label{app:ssr}
\begin{figure}[H]
    \centering
    \includegraphics[width=\textwidth]{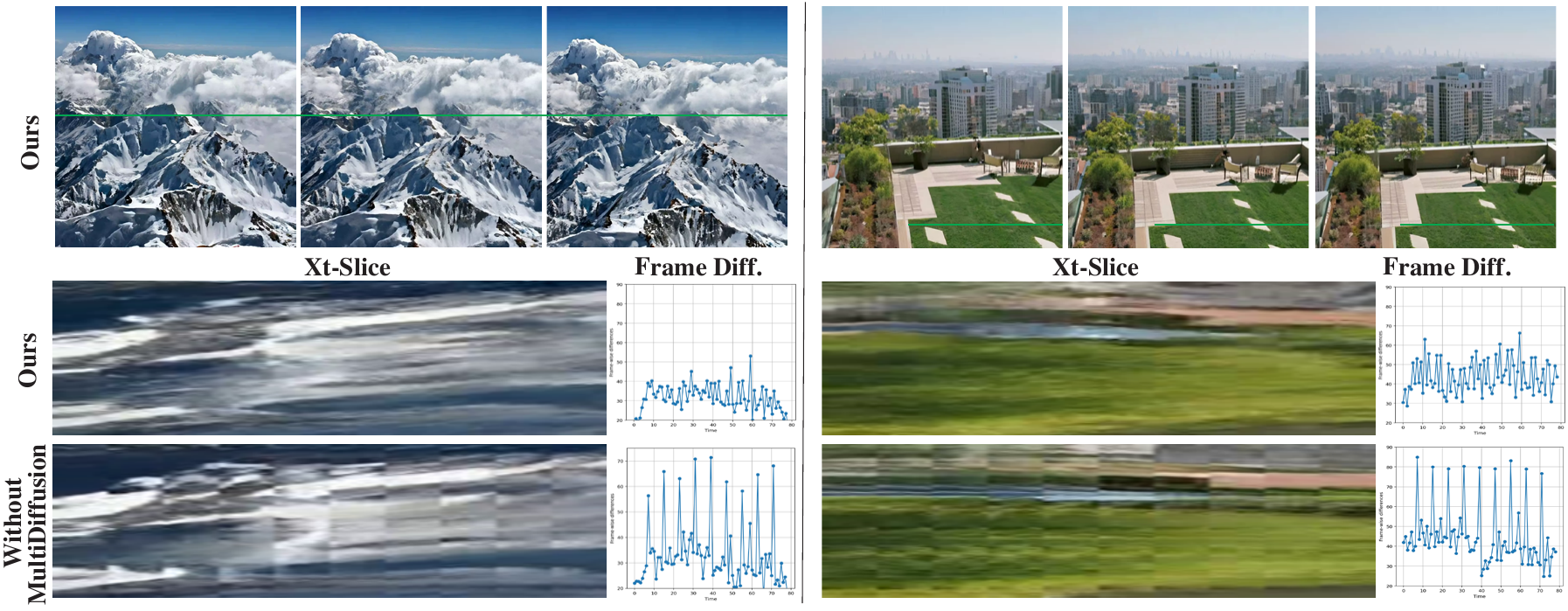}
    \caption{\textbf{MultiDiffusion Ablation.} We ablate the use of MultiDiffusion~\cite{bar2023multidiffusion} for the SSR network. We show sample high-resolution SSR samples in the first row. Below, we show xt-slice of our result and of of the SSR network applied without MultiDiffusion. As can be seen, without MultiDiffusion, there are temporal boundary artifcats. We also demonstarte this property by computing Frame Diff. Specifically, we compute the squared difference between adjacent frames. Without MultiDiffusion, there is a high peak of error between every non-overlapping segment. See App.~\ref{app:ssr}.}
    \label{fig:multi_ablate}
\end{figure}

As discussed in Sec.~\ref{sec:multidiffusion}, we apply MultiDiffusion~\cite{bar2023multidiffusion} along the temporal axis in order to feed segments of the video to the SSR network. Specifically, the temporal windows have an overlap of 2 frames, and at each denoising step we average the predictions of overlapping pixels. MultiDiffusion allows us to avoid temporal boundary artifacts between segments of the video (see Fig.~\ref{fig:multi_ablate}).

\section{Evaluation Protocol}
\label{app:eval}

\subsection{User study}\label{sec:app_user_study}
\begin{figure}[H]
    \centering
    \includegraphics[width=0.6\textwidth]{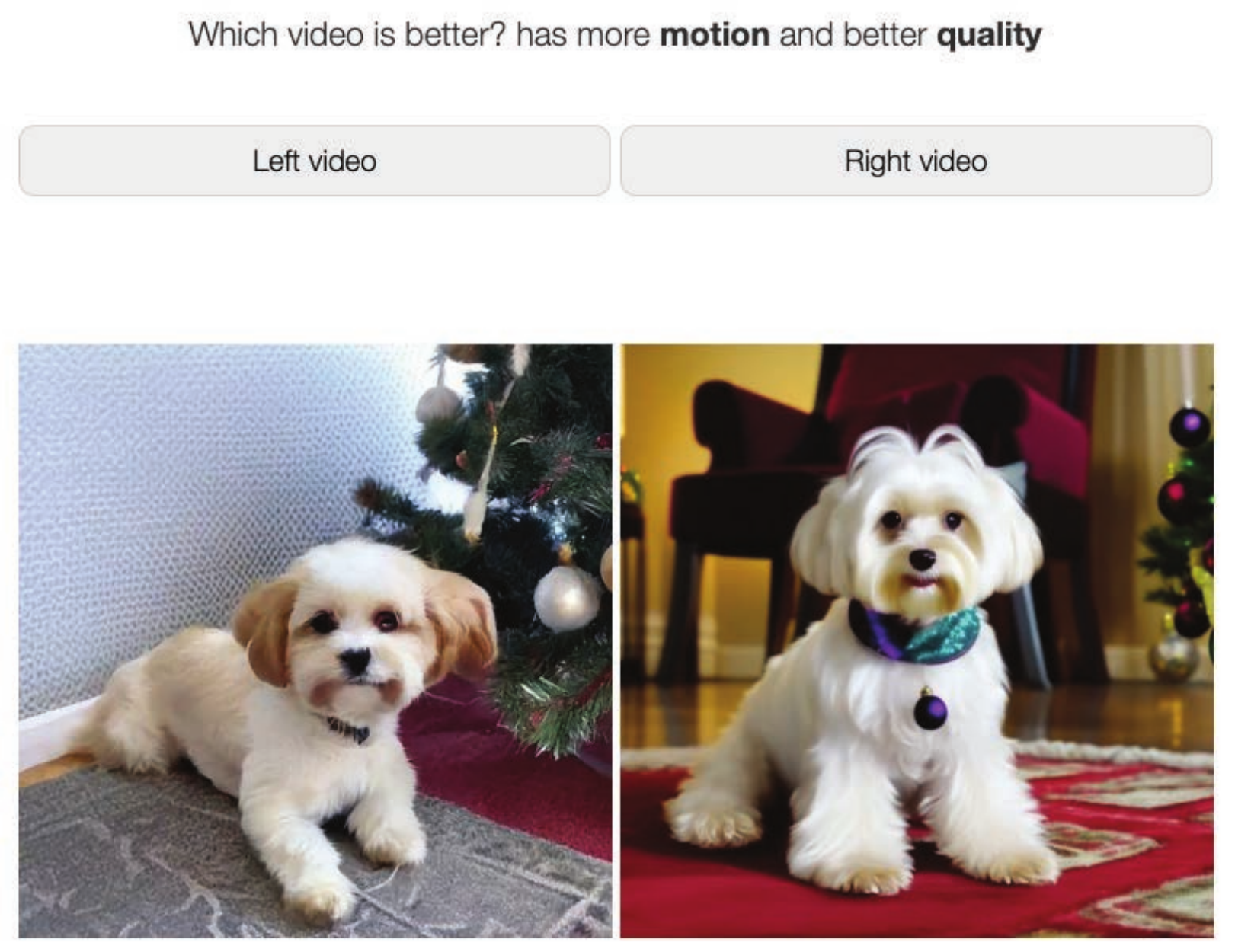}
    \caption{\textbf{User study display.} The user is shown
    two videos, ours and a baseline, randomly ordered.}
    \label{fig:user_study_screen}
\end{figure}

We used a set of 109 text prompts containing a variety of objects and actions for the text-to-video and image-to-video surveys. $91$ prompts were taken from various recent text-to-video generation methods \cite{singer2022make,ho2022imagen,blattmann2023videoldm}. The rest are prompts we created which describe complex scenes and actions, see~\cref{sec:list_of_prompts} for the full list of prompts. For each baseline method, we generated results for all prompts through their official APIs. 
To ensure a fair comparison, when generating results using our method we've fixed a random seed and generated all results using the same random seed without any curation. 
In addition, following \cite{girdhar2023emu} we align the spatial resolution and temporal duration of our videos to each one of the baselines. Spatially, we apply a central crop followed by resizing to 512$\times$512, and temporally, by trimming the start and the end such that the duration of both videos matches the shortest one.
Each participant is shown 10 side-by-side comparisons between our method and a baseline (randomly ordered). 
We then asked the participant "Which video matches the following text?" and "Which video is better? Has more motion and better quality". We enable the user to answer only after watching the full video, waiting at least 3 seconds, and pressing a button reading ``Ready to answer'' (Fig.~\ref{fig:user_study_screen}).
To ensure credible responses, we include several vigilance tests during the study - a real video vs a complete random noise, and a static image vs a real video.

\subsection{Zero-shot evaluation on UCF101}
To compute Fréchet Video Distance (FVD), we generated 10,235 videos by following the class distribution of the UCF101 dataset. We use the same prompt set from \citet{ge2023preserve} to generate the videos. After resizing all videos to 244$\times$244 resolution, we extract I3D embedding \cite{inception3d} of the first 16 frames of our videos. Then we compute the FVD score between the I3D embedding of ours and that of UCF101 videos.
To compute Inception Score (IS), the same generated videos are used to extract C3D embedding \cite{saito2020train}.

\subsection{List of prompts used for the user study evaluation}\label{sec:list_of_prompts}
\begin{enumerate}[itemsep=1mm, parsep=0mm]
    \item \it{"A bear is giving a presentation in classroom."}
    \item \it{"A bear wearing sunglasses and hosting a talk show."}
    \item \it{"A beautiful sunrise on mars, Curiosity rover. High definition, timelapse, dramatic colors."}
    \item \it{"A beautiful sunrise on mars. High definition, timelapse, dramatic colors."}
    \item \it{"A bicycle on top of a boat."}
    \item \it{"A big moon rises on top of Toronto city."}
    \item \it{"A bigfoot walking in the snowstorm."}
    \item \it{"A burning volcano."}
    \item \it{"A car moving slowly on an empty street, rainy evening, van Gogh painting."}
    \item \it{"A cat wearing sunglasses and working as a lifeguard at a pool."}
    \item \it{"A confused grizzly bear in calculus class."}
    \item \it{"A cute happy Corgi playing in park, sunset, 4k."}
    \item \it{"A dog driving a car on a suburban street wearing funny sunglasses"}
    \item \it{"A dog swimming."}
    \item \it{"A dog wearing a Superhero outfit with red cape flying through the sky."}
    \item \it{"A dog wearing virtual reality goggles in sunset, 4k, high resolution."}
    \item \it{"A fantasy landscape, trending on artstation, 4k, high resolution."}
    \item \it{"A fat rabbit wearing a purple robe walking through a fantasy landscape."}
    \item \it{"A fire dragon breathing, trending on artstation, slow motion."}
    \item \it{"A fox dressed in suit dancing in park."}
    \item \it{"A giraffe underneath a microwave."}
    \item \it{"A glass bead falling into water with a huge splash. Sunset in the background."}
    \item \it{"A golden retriever has a picnic on a beautiful tropical beach at sunset, high resolution."}
    \item \it{"A goldendoodle playing in a park by a lake."}
    \item \it{"A hand lifts a cup."}
    \item \it{"A happy elephant wearing a birthday hat walking under the sea."}
    \item \it{"A horse galloping through van Gogh's Starry Night."}
    \item \it{"A hot air balloon ascending over a lush valley, 4k, high definition."}
    \item \it{"A koala bear playing piano in the forest."}
    \item \it{"A lion standing on a surfboard in the ocean in sunset, 4k, high resolution."}
    \item \it{"A monkey is playing bass guitar, stage background, 4k, high resolution."}
    \item \it{"A panda standing on a surfboard in the ocean in sunset, 4k, high resolution."}
    \item \it{"A panda taking a selfie."}
    \item \it{"A polar bear is playing bass guitar in snow, 4k, high resolution."}
    \item \it{"A raccoon dressed in suit playing the trumpet, stage background, 4k, high resolution."}
    \item \it{"Red sports car coming around a bend in a mountain road."}
    \item \it{"A shark swimming in clear Carribean ocean."}
    \item \it{"A sheep to the right of a wine glass."}
    \item \it{"A shiny golden waterfall flowing through glacier at night."}
    \item \it{"A shooting star flying through the night sky over mountains."}
    \item \it{"A sloth playing video games and beating all the high scores."}
    \item \it{"A small hand-crafted wooden boat taking off to space."}
    \item \it{"A squirrel eating a burger."}
    \item \it{"A stunning aerial drone footage time lapse of El Capitan in Yosemite National Park at sunset."}
    \item \it{"A swarm of bees flying around their hive."}
    \item \it{"A teddy bear is playing the electric guitar, high definition, 4k."}
    \item \it{"A teddy bear running in New York City."}
    \item \it{"A teddy bear skating in Times Square."}
    \item \it{"A video of the Earth rotating in space."}
    \item \it{"A video that showcases the beauty of nature, from mountains and waterfalls to forests and oceans."}
    \item \it{"Aerial view of a hiker man standing on a mountain peak."}
    \item \it{"Aerial view of a lake in Switzerland"}
    \item \it{"Aerial view of a snow-covered mountain."}
    \item \it{"Aerial view of a white sandy beach on the shores of a beautiful sea, 4k, high resolution."}
    \item \it{"An animated painting of fluffy white clouds moving in sky."}
    \item \it{"An astronaut cooking with a pan and fire in the kitchen, high definition, 4k."}
    \item \it{"An astronaut riding a horse, high definition, 4k."}
    \item \it{"An astronaut riding a horse."}
    \item \it{"An astronaut riding a horse in sunset, 4k, high resolution."}
    \item \it{"Artistic silhouette of a wolf against a twilight sky, embodying the spirit of the wilderness.."}
    \item \it{"Back view on young woman dressed in a yellow jacket walking in the forest."}
    \item \it{"Maltipoo dog on the carpet next to a Christmas tree in a beautiful living room."}
    \item \it{"Beer pouring into glass, low angle video shot."}
    \item \it{"Bird-eye view of a highway in Los Angeles."}
    \item \it{"Campfire at night in a snowy forest with starry sky in the background."}
    \item \it{"Cartoon character dancing."}
    \item \it{"Cherry blossoms swing in front of ocean view, 4k, high resolution."}
    \item \it{"Chocolate syrup pouring on a vanilla ice cream"}
    \item \it{"Close up of grapes on a rotating table. High definition."}
    \item \it{"Teddy bear surfer rides the wave in the tropics."}
    \item \it{"Drone flythrough interior of sagrada familia cathedral."}
    \item \it{"Drone flythrough of a fast food restaurant on a dystopian alien planet."}
    \item \it{"Drone flythrough of a tropical jungle covered in snow."}
    \item \it{"Edge lit hamster writing code on a computer in front of a scifi steampunk machine with yellow green red and blue gears high quality dslr professional photograph"}
    \item \it{"Epic tracking shot of a Powerful silverback gorilla male walking gracefully",}
    \item \it{"Fireworks."}
    \item \it{"Flying through a temple in ruins, epic, forest overgrown, columns cinematic, detailed, atmospheric, epic, mist, photo-realistic, concept art, volumetric light, cinematic epic, 8k"}
    \item \it{"Flying through an intense battle between pirate ships in a stormy ocean."}
    \item \it{"Flying through fantasy landscapes, 4k, high resolution."}
    \item \it{"Pug dog listening to music with big headphones."}
    \item \it{"Horror house living room interior overview design, 8K, ultra wide angle, pincushion lens effect."}
    \item \it{"Incredibly detailed science fiction scene set on an alien planet, view of a marketplace. Pixel art."}

    \item \it{"Jack russell terrier dog snowboarding. GoPro shot."}
    \item \it{"Low angle of pouring beer into a glass cup."}
    \item \it{"Melting pistachio ice cream dripping down the cone."} 
    \item \it{"Milk dripping into a cup of coffee, high definition, 4k."}
    \item \it{"Pouring latte art into a silver cup with a golden spoon next to it.}
    \item \it{"Shoveling snow."}
    \item \it{"Studio shot of minimal kinetic sculpture made from thin wire shaped like a bird on white background."}
    \item \it{"Sunset time lapse at the beach with moving clouds and colors in the sky, 4k, high resolution."}
    \item \it{"Teddy bear walking down 5th Avenue, front view, beautiful sunset, close up, high definition, 4k."}
    \item \it{"The Orient Express driving through a fantasy landscape, animated oil on canvas."}
    \item \it{"Time lapse at a fantasy landscape, 4k, high resolution."}
    \item \it{"Time lapse at the snow land with aurora in the sky, 4k, high resolution."}
    \item \it{"Tiny plant sprout coming out of the ground."}
    \item \it{"Toy Poodle dog rides a penny board outdoors"}
    \item \it{"Traveler walking alone in the misty forest at sunset."}
    \item \it{"Turtle swimming in ocean."}
    \item \it{"Two pandas discussing an academic paper."}
    \item \it{"Two pandas sitting at a table playing cards, 4k, high resolution."}
    \item \it{"Two raccoons are playing drum kit in NYC Times Square, 4k, high resolution."}
    \item \it{"Two raccoons reading books in NYC Times Square."}
    \item \it{"US flag waving on massive sunrise clouds in bad weather. High quality, dynamic, detailed, cinematic lighting"}
    \item \it{"View of a castle with fantastically high towers reaching into the clouds in a hilly forest at dawn."}
    \item \it{"Waves crashing against a lone lighthouse, ominous lighting."}
    \item \it{"Woman in white dress waving on top of a mountain."}
    \item \it{"Wood on fire."}
    \item \it{"Wooden figurine surfing on a surfboard in space."}
    \item \it{"Yellow flowers swing in wind."}
\end{enumerate}




\end{document}